\documentclass[12pt,twoside,a4paper]{article}
\usepackage{amsmath,amssymb,latexsym,theorem,natbib,epsfig,color,subfigure,footmisc,bbm}
\usepackage{multirow,graphicx,array,rotating,url,multicol, booktabs, pifont}  
\usepackage{adjustbox}
\usepackage{epstopdf,afterpage,wasysym,hyperref}
\usepackage{verbatim}
\usepackage{amssymb}
\usepackage[normalem]{ulem}
\newcommand{\cmark}{\ding{51}}
\newcommand{\xmark}{--}

\title{\Large Statistical versus machine learning-based spatial interpolation of post-processed ensemble weather forecasts}

\author{Mária Lakatos\\
{\small Faculty of Informatics, University of Debrecen, Hungary}
}

\date{}

\begin{document}
\maketitle

\begin{abstract}
Statistical post-processing improves ensemble weather forecasts, but generating calibrated predictions at locations without observations remains challenging. This study compares statistical and machine-learning-based methods for post-processing ECMWF 2-m temperature and 10-m wind speed forecasts at observed and unobserved stations in Germany. We consider EMOS-based approaches, distributional regression networks, Transformers, and graph neural networks under both limited and extended predictor settings. For temperature, we also investigate linear forecast combinations and propose an altitude-aware linear pool (ALP). The results show that post-processing improves upon the raw ensemble in most settings, but no single method performs best across all variables, station groups, and evaluation metrics. The proposed ALP provides a small but significant improvement over the standard linear pool at unobserved locations.

\end{abstract}

\section{Introduction}
\label{sec1}
Representing a major step forward in estimating the uncertainty of future events, the introduction of ensemble prediction systems (EPSs) in operational use marked a significant milestone in advancing meteorology. Today, as one of the world's leading weather centres, the European Centre for Medium-Range Weather Forecasts (ECMWF) operates both the physics-based Integrated Forecasting System \citep[IFS;][]{ECMWF2026IFSPartV} and the data-driven Artificial Intelligence (AI) Forecasting System \citep[AIFS;][]{aifs-crps26}, each providing operational ensemble forecasts. However, forecasts produced by both systems often exhibit systematic errors, which can lead to biased and poorly calibrated predictions \citep{gneiting2026, kocsis2026}, necessitating statistical post-processing.

Numerous techniques now exist to improve these predictions, ranging from classical statistical methods to state-of-the-art AI models \citep{vbd21}. These methods could differ in their underlying assumptions, model complexity, and the way they exploit predictor information to estimate the predictive distribution. One of the most widely used statistical post-processing methods is ensemble model output statistics \citep[EMOS;][]{grwg05}, also known as non-homogeneous regression, which assumes a linear relationship between the predictors and the predictand. In contrast, AI-based \citep{rl18, chen2024} and other non-parametric approaches can capture complex nonlinear relationships and flexibly incorporate diverse predictor variables \citep{Muggeo2013, b20bqn}. However, these methods generally require past observations for training and therefore cannot produce calibrated forecasts for locations and time points where such data are unavailable.

In order to ensure that calibration can be carried out under such conditions, several methods have been proposed that can be used for spatial interpolation of calibrated predictions. For example, for temperature, \cite{scheuerer2014spatially} proposed a spatial extension of EMOS in which the local temperature averages and forecast uncertainty parameters are modeled by Gaussian random fields and interpolated to arbitrary locations, while \cite{baran2024clustering} propose a general clustering-based interpolation technique of extending calibrated predictive distributions from observation stations to any location in the ensemble domain where there are ensemble forecasts at hand.

While existing approaches rely primarily on geostatistical techniques to extend already calibrated forecasts to unobserved locations  in a second step \citep[see e.g.,][]{Kleiber2011}, data-driven models provide an alternative by learning the spatial mapping directly from raw ensemble forecasts and observations. Different neural network architectures incorporate different inductive biases, which determine how spatial relationships are represented and learned \citep{battaglia2018relational}. As a result, their ability to generalize to previously unseen locations may differ, potentially resulting in differences in forecast quality. More importantly, the generalization ability of neural networks allows post-processing and ``spatial interpolation'' to be performed jointly within a single end-to-end framework, in contrast to the two-step approaches of \cite{baran2024clustering} and \cite{scheuerer2014spatially}.

Motivated by these properties, this study investigates whether three neural network-based approaches %architectures 
(distributional regression network, %multilayer perceptron,
graph neural network, and transformer) can perform statistical post-processing of raw ECMWF temperature and wind speed ensemble forecasts while simultaneously providing predictions at locations without historical observations. More importantly, our goal is to examine how the performance of these single-step AI models compares with the general clustering-based statistical interpolation method of \cite{baran2024clustering}, in which the EMOS approach was applied for initial calibration at observed locations. To ensure a fair comparison, in our case studies we employed two feature sets for each model: one consisting solely of explanatory variables used in the standard EMOS framework (specifically, the mean and standard deviation %and variance 
of the raw ensemble forecasts), and another utilizing a wide range of additional covariates. For the latter, we used a boosted EMOS model incorporating automatic variable selection as the reference \citep{mmz17}, replacing the standard EMOS. To the best of the authors' knowledge, no previous study has systematically compared end-to-end deep learning models with statistical post-processing methods for the task of spatial interpolation of post-processed ensemble forecasts.  In addition, for temperature forecasts, we also investigate the predictive skill of different 
versions of linear pool \citep{gneiting2013combining} to combine predictive distributions. Furthermore, we propose an altitude-aware extension of the standard linear pool. In this approach, the combination weights vary with station altitude, and the best-performing EMOS variant is combined with each of the machine learning-based models.

The remainder of the paper is organized as follows. Section 2 introduces the ECMWF temperature and wind speed data used in our case studies. Section 3 presents the normal and truncated normal EMOS models, their boosted variants, the machine-learning models, and the linear pools, followed by the interpolation approaches and model verification procedure. The results are reported in Section 4, followed by a discussion in Section 5.

\section{Data}
\label{sec2}

\begin{figure}
    \centering
    \includegraphics[width=.8\linewidth]{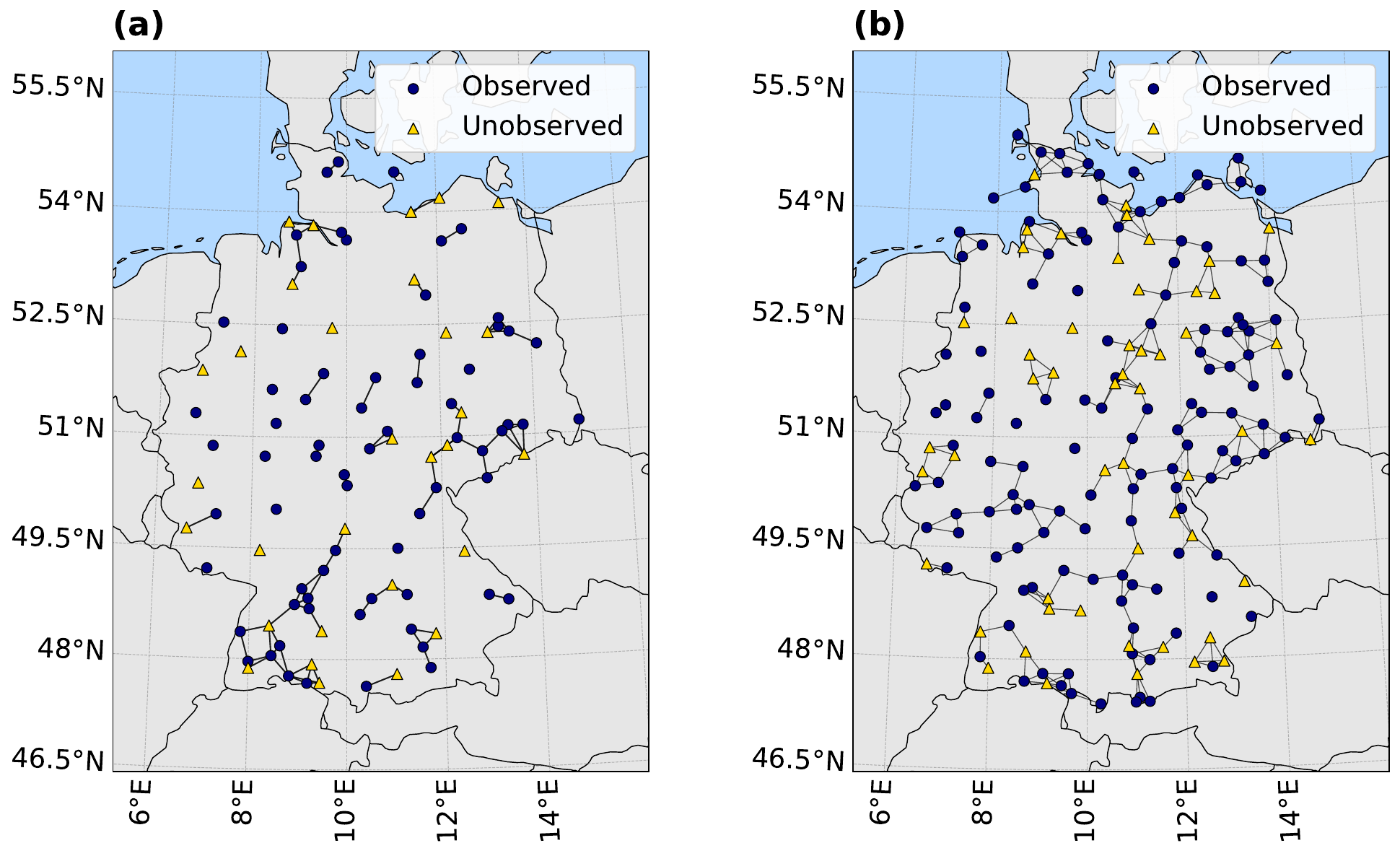}
    \caption{Locations of (a) temperature and (b) wind speed observation stations across Germany. Black lines indicate edges used to construct the input graph for the GraphSAGE models.}
    \label{fig:maps_with_graphs}
\end{figure}

In this study, we utilize 50-member ECMWF ensemble forecasts of 2-m temperature (T2M) and 10-m wind speed (WS) for Germany with a lead time of 48 hours from the dataset published by \cite{chen2024}. 
Ensemble forecasts are issued once per day at 00 UTC and retrieved from the THORPEX Interactive Grand Global Ensemble (TIGGE) archive \citep{bougeault2010thorpex}, which provides forecasts on a $0.5^\circ \times 0.5^\circ$ grid covering Europe. The gridded forecasts are interpolated to the locations of meteorological stations in order to obtain station-level predictor variables. Corresponding observations are obtained from the Climate Data Center of the German Meteorological Service (Deutscher Wetterdienst, DWD). The dataset spans the period from 3 January 2007 to 31 December 2016. From the total of 419 stations with temperature and 199 stations with wind speed data, we restrict the analysis to locations 
for which both observations and forecasts are available for each day in the study period. This results in 99 stations for temperature and 182 stations for wind speed. The temperature dataset contains no missing observations. In contrast, a slightly less restrictive filtering rule is used for wind speed, resulting in a small proportion of missing values (less than 1\%). 

Besides T2M and WS ensemble forecasts, the dataset comprises means and standard deviations of ensemble forecasts of several auxiliary weather quantities, which aligns with the setup of  \citep{rl18}. Further predictors include station metadata such as geographical coordinates, station altitude, and model orography, as well as a sine-transformed representation of the day of the year. A complete list of covariates is provided in Table 1 of \cite{chen2024}, and is also included as Table \ref{tab:predictors} in the Appendix of the present paper for convenience. The final predictor set contains 41 variables for T2M and 46 variables for WS,  70\% of the stations were treated as observed stations and used for training, corresponding to 69 stations for T2M and 129 stations for WS. The remaining 30\% of the stations were reserved for testing, where the interpolation performance of the considered methods was evaluated.

\section{Post-processing methods and forecast verification}
\label{sec3}

In the following sections, let us denote by \ $f_1,f_2, \ldots ,f_{50}$ \ the 50-member operational ECMWF ensemble forecast of a given weather quantity for a given location, time, and forecast horizon, and denote by \ $\overline f$ \ and \ $S^2$ \ the ensemble mean and variance, respectively; that is
\begin{equation*}
  \overline f:=\frac 1{50}\sum_{k=1}^{50}f_k \qquad
  \text{and} \qquad S^2:=\frac 1{49}\sum_{k=1}^{50} \big(f_k - \overline f\big)^2.
\end{equation*}

\subsection{Benchmark univariate statistical post-processing approaches}
\label{subs3.1}

As mentioned in the Introduction, %gondolom lesz benne valami ilyen
over the past decades a wide range of methods have been developed for post-processing ensemble forecasts of a single variable for a given location, time, and lead time. 

%Nem tudom, mit akarsz még ide tenni, de nyilván annak megfelelően kell változtatni a fejezet struktúrát.
\subsubsection{Ensemble model output statistics}
\label{subs3.1.1}

Ensemble model output statistics \citep[EMOS;][]{grwg05}, or nonhomogeneous regression, is a simple parametric post-processing method, that provides a calibrated probabilistic forecast in the form of a full predictive distribution. %Lehet, hogy ez megint szerepel valahol korábban
In general, the conditional distribution of the future value \ $Y$ \ of a weather quantity with respect to the corresponding 50-member ensemble forecast is expressed as
\begin{equation*}
  Y \mid f_1, f_2, \ldots ,f_{50} \sim h\big (y\mid f_1, f_2, \ldots ,f_{50}; \theta \big),
\end{equation*}
where the distributional parameters of the conditional probability density function (PDF) \ $h\big (y\mid f_1, f_2, \ldots ,f_{50}; \theta \big)$ \ are connected to the forecast ensemble via suitable link functions depending on the parameter (vector) \ $\theta$. \ EMOS models for various predictable quantities differ in the considered parametric distribution family and/or in the applied link functions, where, following the suggestions of \citet{gneiting14} \citep[see also][Section 8.3.2]{w26}, exchangeable ensemble members are weighted equally.

For temperature forecasts, the normal distribution with affine link functions is the most popular choice \citep{grwg05}. In particular, for the 50-member ECMWF ensemble, the mean \ $\mu$ \ and standard deviation \ $\sigma>0$ \ of the Gaussian predictive distribution \ ${\mathcal N}\big(\mu,\sigma^2\big)$ \ are given as
\begin{equation}
  \label{eq:emosLink}
  \mu = a + b^2 \overline f \qquad \text{and} \qquad \sigma = c^2+d^2S.
\end{equation}
Model parameters \ $a,b,c,d \in {\mathbb R}$ \ are estimated following the optimum score principle of \citet{gr07}, that is, by minimizing the mean of a proper scoring rule over the training data comprising past forecast-observation pairs. The usual choice is the continuous ranked probability score \citep[CRPS;][Section 9.5.1]{w26}, described in detail in Section \ref{subs3.5}.

In contrast to temperature, wind speed requires a right-skewed distribution with a positive support, such as a truncated normal (TN) distribution with left truncation at zero \citep{tg10}, log-normal distribution \citep{bl15}, generalized extreme value distribution \citep{lt13} or its left-truncated version \citep{bszsz21}. Here we consider the TN EMOS model of \citep{tg10} with link functions \eqref{eq:emosLink}, where now \ $\mu$ \ and \ $\sigma$ \ stand for the location and scale of the left-truncated normal law \ ${\mathcal N}_0\big(\mu,\sigma^2\big)$, \ respectively.

\subsubsection{Boosted ensemble model output statistics}
\label{subs3.1.2}

A general disadvantage of EMOS models is that, in contrast to the machine learning-based parametric distributional regression network \citep[DRN;][]{rl18} approach, %erről persze lehet lesz szó korábban
the rigid structure of the link functions does not allow an easy accommodation of additional model covariates such as forecasts of related quantities or location- or time-specific information. In principle, one can include such predictors linearly both in the location and the scale parameter; however, a large number of covariates might easily result in overfitting, which has an inferior effect on forecast performance. These problems can be bypassed by the boosting algorithm of \citet{mmz17}, which, under the assumption of affine link functions, performs automatic feature selection and provides the parameter estimates for the obtained models. For the considered normal and truncated normal predictive distributions, we utilize the implementation provided in the {\tt R} package {\tt crch} \citep{mmz16}.%; in what follows, the corresponding boosted EMOS models are referred to as EMOS-B. 

%\subsection{Deep learning post-processing models} 

\subsection{Machine-learning-based post-processing models}
\label{subs3.2}
This section briefly introduces the machine-learning (ML) models used in this study. Similar to the EMOS model described in Section \ref{subs3.1.1}, each model predicts the location and scale parameters of either a Gaussian law or a Gaussian law left-truncated at zero by minimizing the corresponding closed-form CRPS. Here we consider three popular approaches based on different network architectures. The first is a simple feed-forward multilayer perceptron (MLP) neural network, which served as the basic architecture of the distributional regression network (DRN) model of \citet{rl18}. Then we utilize the more sophisticated graph neural networks (GNNs) and transformer architectures that are explicitly tailored for spatial modelling.
The main hyperparameter settings are summarized in Appendix Table \ref{tab:ml_hyperparameters}. %All machine-learning models are trained in a rolling setting using 3283 training days, and the hyperparameters are selected by a grid-style based search. %Each experiment is repeated for 10 random seeds to reduce the effect of random initialization. Input features are standardized within each rolling training period using only the training data.

%\subsubsection{Multilayer perceptron}
\subsubsection{Distributional regression network}
\label{subs3.2.1}

Distributional regression networks (DRNs) were first applied to ensemble
forecast post-processing by \cite{rl18}. A DRN uses a multilayer perceptron
(MLP) to link the input features to the parameters of a predictive
distribution. The input features pass through one or more fully connected
hidden layers, while the output layer provides the parameters of the
calibrated predictive distribution. DRNs can include additional predictors
that are difficult to incorporate into standard EMOS models and may therefore
outperform EMOS when informative covariates are available. Since their
introduction, machine-learning methods have been applied to both continuous
and discrete weather variables, including visibility and total cloud cover
\citep{lakatos2024enhancing, baran2021machine}. For further details on MLPs,
we refer to \cite{gbc16}.

\subsubsection{Graph neural network}
\label{subs3.2.2}

Graph neural networks (GNNs) are designed for spatial applications and problems in which the relationships between nodes are irregular, such as weather stations that are unevenly distributed across a region. In the case of post-processing ensemble forecasts, observation stations represent the nodes of the graph and the edges define the connections between them \citep{feik2024graph}. These connections can be based on geographical distance or a fixed number of nearest neighboring stations. Each node is associated with an $n$-dimensional feature vector, which may include relevant covariates, measurements, or summary statistics. By propagating and aggregating information along the graph edges, GNNs can capture both local and global dependencies \citep{lakatos2026}. When several GNN layers are used, information can also be collected from more distant nodes. In this study, the GraphSAGE architecture is employed \citep{hamilton2017inductive}, using mean aggregation to combine information from neighboring stations. In our implementation, the graph is constructed from the fixed station coordinates using the Haversine distance. Two stations are connected if their distance is at most 50 km, resulting in an undirected binary graph that remains fixed over all rolling training periods. No explicit self-loops are added, and stations without another station within 50 km may remain isolated (see Figure \ref{fig:maps_with_graphs}).

\subsubsection{Transformer}
\label{subs3.2.3}
Unlike GNNs, which collect information from predefined neighboring nodes,
Transformers use self-attention to connect each location with every other
location \citep{poecke25}. In self-attention, each target station is compared
with all other stations, and the resulting similarity scores determine how
strongly the information from each station contributes to the prediction at
the target location. During training, the model learns these relationships
between stations. Several attention heads work in parallel and can capture
different types of relationships. In this way, the model can represent both
nearby and long-range spatial dependencies without relying on a predefined
graph. For more details about the Transformer architecture, we refer to
\cite{vaswani2017attention}.

\subsection{Standard- and altitude-aware linear pools} 
\label{subs3.3}

The standard linear pool (SLP) has been used in several studies to improve forecast
performance, although its application has usually been limited to
observed locations \citep{gneiting2013combining, bassetti2018bayesian, baran2018combining}. In Section \ref{sec4}, we extend this approach to
unobserved stations in the case when the component predictive distributions are both Gaussian.
Let \(F_{d,s}^{(w)}\) denote the CDF corresponding to the mixture for
day \(d\) and station \(s\), that is,

\[
F_{d,s}^{(w)}(y)
=
w\Phi\big(y - \mu_{1,d,s})/\sigma_{1,d,s}\big)
+
(1-w)\Phi\big((y - \mu_{2,d,s})/\sigma_{2,d,s}\big),
\]
where $\Phi(y)$ denotes the standard Gaussian CDF, \  \(\mu_{k,d,s} \text{ and } \sigma_{k,d,s},\) \ \(k=1,2\), are the location and scale parameters of the two Gaussian components, respectively, and \(w\in[0,1]\) is the mixture weight.
The weights are estimated by minimizing the mean CRPS of the pooled mixture distribution over a rolling training window at the observed stations,
which estimates are then applied to unobserved stations. Note that for the Gaussian mixture the CRPS can be expressed in a closed form \citep{jkl19}. Let \(\mathcal{W}_t\) denote the rolling training window before verification
day \(t\), \(\mathcal{S}_{\mathrm{obs}}\) be the set of observed
stations, and let \(y_{d,s}\) the observation at station \(s\) on day \(d\). The SLP weight is estimated as
\[
\widehat{w}_t
=
\underset{w\in[0,1]}{\arg\min}
\frac{1}{|\mathcal{W}_t||\mathcal{S}_{\mathrm{obs}}|}
\sum_{d\in\mathcal{W}_t}
\sum_{s\in\mathcal{S}_{\mathrm{obs}}}
\operatorname{CRPS}\!\left(F_{d,s}^{(w)},y_{d,s}\right).
\]
The resulting estimate \(\widehat{w}_t\) is then applied to both observed
and unobserved stations on day \(t\). However, one global weight may not be suitable for stations at very
different altitudes. We therefore introduce an altitude-aware linear pool (ALP)
as a simple extension of the standard approach. For the ALP, the stations are divided into three altitude groups:
low (\(<300\) m), medium (\(300\)--\(700\) m), and high
(\(\geq700\) m). Let \(g(s)\in\{1,2,3\}\) denote the altitude group of
station \(s\). Instead of estimating one global weight
\(\widehat{w}_t\), the same CRPS minimization is performed separately
using the observed stations within each altitude group, resulting in
group-specific weights \(\widehat{w}_{t,g}\). The predictive distribution
at station \(s\) is then
\[
F_{\mathrm{ALP},t,s}(y)
=
\widehat{w}_{t,g(s)}F_{1,t,s}(y)
+
\left(1-\widehat{w}_{t,g(s)}\right)F_{2,t,s}(y).
\]
Thus, each unobserved station receives the weight estimated for its
corresponding altitude group.

\subsection{Training data selection and spatial interpolation}
\label{subs3.4}

Temporal and spatial selection of training data, which typically consist of past forecast-observation pairs and possibly additional covariates, plays a crucial role in the efficiency of various post-processing methods \citep{llmssz20}. In our case studies, we use rolling training periods, meaning that the models are trained using forecast--observation pairs from the \(n\) preceding days available before the test day.

%In what follows, we describe different spatial selection methods for constructing the training data and briefly the applied spatial interpolation methods.

%When post-processing ensemble forecasts, the training data typically consist of past forecast-observation pairs and possibly additional covariates. One 
In terms of spatial selection, one option is to use all available historical data from all stations. This approach, often referred to as regional, provides a large training dataset and can lead to more stable parameter estimates \citep{tg10}. However, it may ignore local differences between stations and therefore perform less well in regions with specific geographical or climatic conditions. Another approach, referred to as local approach, uses only historical forecast--observation pairs from the station whose forecasts are being calibrated. Its main advantage is that it can capture station-specific characteristics. However, it requires a longer training period for stable estimation. Finally, the semi-local method of \cite{lerch2017similarity} uses \(k\)-means clustering to group stations with similar climatological conditions or forecast errors, and estimates a common set of parameters for all stations within each group, thereby combining the advantages of the local and regional approaches. In particular, we follow the "forecasts as features" setup of \citet{baran2024clustering}. This means that at each modelling step to each location we assign a feature vector comprising the concatenation of the scaled training vectors of the input meteorological variables (for instance, ensemble mean and standard deviation for EMOS), which vectors are then used for clustering the observed stations. Note that in the rolling training period setup, for each date in the verification period a new set of clusters is formed dynamically.  In the case studies in Section \ref{sec4}, the EMOS and boosted EMOS models use all three spatial selection strategies. We use the notation EMOS-R for regional, EMOS-L for local, and EMOS-C for semi-local modelling. Since the neural network models use all available historical data for training, their strategy is equivalent to regional modelling in the EMOS framework. 

In our case studies, the neural networks use a simple interpolation approach. Only data from observed stations are used for training, after which the networks provide forecasts for all stations based on the available input features. Therefore, they can naturally generate predictions for locations not used during training. In the GNN, stations are connected based on geographical proximity, allowing information from neighboring observed stations to be propagated to unobserved locations. For the Transformer, information is shared between stations through the attention mechanism, which allows the model to learn relationships between observed and unobserved locations from their available input features. Thus, all models can generate forecasts at stations whose observations were excluded from training.

Similarly, for regional EMOS models (EMOS-R), at unobserved locations, one uses the single set of model parameters valid for the actual test date for all observed stations. Furthermore, for EMOS-C, we follow the approach of \citet{baran2024clustering} by assigning each unobserved location to the most similar cluster in terms of the Euclidean distance of the feature vector of the given unobserved station (calculated as given above) to the cluster mean. Then the unobserved location inherits the EMOS parameters valid for the assigned cluster. Finally, EMOS-L can be considered as a special case of EMOS-C when each observed station forms an individual cluster.

% mentioned earlier, to obtain EMOS forecasts at unobserved locations, we follow the approach of \cite{baran2024clustering}. This method extends the semi-local model of \cite{lerch2017similarity}. First, stations are grouped according to similarities in their raw forecasts, and the EMOS parameters are then estimated separately for each group. Unobserved stations inherit the parameters of the group to which they are assigned, while in the regional case, all unobserved locations use the same global set of EMOS coefficients. 

%The temporal construction of the training data is another important aspect. In our case studies, we use rolling training periods, meaning that the models are trained using forecast--observation pairs from the \(n\) preceding days available before the test day.

\subsection{Forecast verification} 
\label{subs3.5}

The aim of probabilistic forecasting is to produce predictions that are both sharp and well calibrated. One of the most widely used measures that considers both aspects in atmospheric science is the CRPS.
It is a proper scoring rule that measures how well the predictive distribution agrees with the observed value. Let \(F\) denote the cumulative distribution function of the predictive
distribution, and let \(y\) be the corresponding observation. The CRPS is
defined as
\[
\operatorname{CRPS}(F,y)
:=
\int_{-\infty}^{\infty}
\left(
F(x)-\mathbb{I}\{x\geq y\}
\right)^2
\,\mathrm{d}x,
\]
where \(\mathbb{I}\{\cdot\}\) denotes the indicator function. 

Another common tool for assessing probabilistic calibration is the verification rank histogram \citep[see e.g.][Section 9.7.1]{w26}. For a \(K\)-member ensemble forecast, the verification rank is the rank of the observation among the
ordered ensemble members. Thus, the verification rank can take integer values from \(1\) to \(K+1\).
In this study, \(K=50\), so there are \(51\) possible ranks. A uniform verification rank histogram indicates a
calibrated ensemble. The deviation from uniformity can be measured by
the reliability index \citep[RI;][]{dmhzds06}
\[
\operatorname{RI}
:=
\sum_{r=1}^{K+1}
\left|
\rho_r-\frac{1}{K+1}
\right|,
\]
where \(\rho_r\) is the relative frequency of rank \(r\).

\begin{table}[t]
  \centering

  {\scriptsize
  \setlength{\tabcolsep}{5pt}

  \begin{tabular}{lccccc}
    \toprule
    \multicolumn{6}{c}{\textbf{Non-extended feature set}} \\
    \cmidrule(lr){1-6}
    & EMOS-R & EMOS-C & EMOS-L & TR & GNN-Geo \\
    \midrule
    Observed
    & 1.436$^{*}$ & 3.270$^{*}$ & \textbf{14.710$^{*}$}
    & 0.419$^{*}$ & 4.420$^{*}$ \\
    Unobserved
    & \textbf{2.515$^{*}$} & -0.234 & -4.454
    & 0.659$^{*}$ & -1.951 \\
    \bottomrule
  \end{tabular}
  \begin{tabular}{lccccc}
    \toprule
    \multicolumn{6}{c}{\textbf{Extended feature set}} \\
    \cmidrule(lr){1-6}
    & EMOS-R & EMOS-C & EMOS-L & TR & GNN-Geo \\
    \midrule
    Observed
    & -13.130 & -9.697 & 0.953
    & \textbf{3.841$^{*}$} & 2.661$^{*}$ \\
    Unobserved
    & -1.717 & -5.164 & -21.878
    & \textbf{0.614} & -8.092 \\
    \bottomrule
  \end{tabular}
  }

  \caption{Group-wise CRPS skill scores of T2M forecasts relative to the
  corresponding DRN baseline. Positive values indicate improvement, while
  bold values mark the best model within each row and feature set. Asterisks
  indicate significant improvement according to one-sided Diebold--Mariano
  tests at the 5\% level.}
  \label{tab:crpss_table_t2m}
\end{table}

Finally, we also assess calibration by reporting the coverages of central prediction intervals. For a given confidence level $\alpha \in (0,1)$, the endpoints of such a $(1-\alpha)\times 100\%$ interval are specified by the $\alpha/2$ and $1-\alpha/2$ quantiles of the predictive distribution, and by coverage we mean the fraction of observations falling within the interval. For a calibrated forecast, this proportion should be around $(1-\alpha)\times 100\%$. To guarantee a fair comparison with the raw ensemble predictions, $\alpha$ is often chosen to match the ensemble nominal coverage, which for a $K$-member ensemble equals $(K-1)/(K+1)\times 100\%$ \ (that is, $\alpha=2/(K+1)$). We also provide the average width of central prediction intervals as a measure of sharpness of the probabilistic forecasts.

When a forecast is represented by a single value, its predictive accuracy is commonly evaluated using the mean absolute error (MAE), which is minimized by the median \citep{GneitingRanjan2011}, and the root mean squared error (RMSE), which is minimized by the mean.

Furthermore, we use the Diebold--Mariano (DM) test to examine whether the
difference between the scores of two forecasts is statistically significant
\citep{diebold1995paring}. The test can account for temporal
dependence between the score differences. Given a scoring rule \(S\) and
two competing probabilistic forecasts \(F\) and \(G\), the DM test statistic
is defined as
\[
t_N
=
\sqrt{N}
\frac{\overline{S}_F-\overline{S}_G}
{\widehat{\sigma}_N},
\]
where \(N\) is the number of forecast cases in the test set,
\(\overline{S}_F\) and \(\overline{S}_G\) are the mean scores of
forecasts \(F\) and \(G\), respectively, and
\(\widehat{\sigma}_N\) is an estimator of the asymptotic standard
deviation of the score differences. Under the null hypothesis of equal predictive
performance, \(t_N\) asymptotically follows a standard Gaussian law.
Since lower scores indicate better forecasts, a negative value of \(t_N\)
favors \(F\), while a positive value favors \(G\). In the case studies of Section \ref{sec4}, the score differences are calculated for each verification
day after averaging the scores over the stations in the corresponding station
group.

Finally, the improvement of a post-processed forecast over a reference
prediction can be measured by skill scores. Let
\(\overline{S}_{\mathrm{model}}\) and
\(\overline{S}_{\mathrm{ref}}\) denote the mean scores of the evaluated
model and the reference forecast, respectively. The skill score is
defined as
\[
\operatorname{SS}
=
1-
\frac{\overline{S}_{\mathrm{model}}}
     {\overline{S}_{\mathrm{ref}}}.
\]
Positive values indicate an
improvement over the reference forecast, while negative values indicate
worse predictive performance. In Section \ref{sec4} we report continuous ranked probability skill score (CRPSS), mean absolute error skill score (MAES), and root mean squared error skill score (RMSES).

\begin{table}[t]
  \centering
  {\scriptsize
  \setlength{\tabcolsep}{5pt}
  \begin{tabular}{lccccc}
    \toprule
    \multicolumn{6}{c}{\textbf{Non-extended feature set}} \\
    \cmidrule(lr){1-6}
    & EMOS-R & EMOS-C & EMOS-L & TR & GNN-Geo \\
    \midrule
    MAES, observed & 0.825 & 2.274$^{*}$ & \textbf{13.380$^{*}$} & 0.136 & 3.719$^{*}$ \\
    MAES, unobserved & \textbf{2.142$^{*}$} & -1.251 & -3.710 & -0.013 & -3.113 \\
    \addlinespace[3pt]
    RMSES, observed & 1.571$^{*}$ & 2.916$^{*}$ & \textbf{13.834$^{*}$} & 0.043 & 3.954$^{*}$ \\
    RMSES, unobserved & \textbf{1.865$^{*}$} & -0.850 & -5.123 & 0.067 & -3.140 \\
    \bottomrule
  \end{tabular}
  \begin{tabular}{lccccc}
    \toprule
    \multicolumn{6}{c}{\textbf{Extended feature set}} \\
    \cmidrule(lr){1-6}
    & EMOS-R & EMOS-C & EMOS-L & TR & GNN-Geo \\
    \midrule
    MAES, observed & -12.733 & -9.223 & 1.651$^{*}$ & \textbf{3.127$^{*}$} & 2.799$^{*}$ \\
    MAES, unobserved & -1.547 & -4.839 & -21.619 & \textbf{0.090} & -7.123 \\
    \addlinespace[3pt]
    RMSES, observed & -13.436 & -9.567 & 1.638$^{*}$ & \textbf{2.956$^{*}$} & 2.698$^{*}$ \\
    RMSES, unobserved & -2.567 & -6.572 & -20.115 & \textbf{-0.157} & -6.877 \\
    \bottomrule
  \end{tabular}
  }
  \caption{Group-wise mean absolute error skill scores (MAES) and root mean squared error skill scores (RMSES) of T2M forecasts relative to the corresponding DRN baseline. Positive values indicate improvement, while bold values mark the best model within each row and feature set. Asterisks indicate significant improvement according to one-sided Diebold--Mariano tests at the 5\% level.}
  \label{tab:mae_rmse_skill_t2m}
\end{table}

\section{Results}
\label{sec4}

This section presents the results of the post-processing of temperature and
wind speed forecasts. As mentioned in the Introduction, we also examine
two-component linear pools combining post-processed EMOS-L and ML forecasts
to assess whether the component forecasts provide complementary information.
The semi-local EMOS and machine-learning models are each fitted 10 times to
account for the random initialization of \(k\)-means clustering and model
weights, respectively. The reported scores are averaged over these runs.

\subsection{Temperature}
\label{sec4.1}

For temperature forecasts, the length of the rolling training period is 3283 days for all machine-learning models, that was selected through a grid-search. For the classical EMOS-R, EMOS-L, and EMOS-C models based on the non-extended feature set, the rolling training periods are 30, 450, and 30 days, respectively, and in the case of EMOS-C, observed stations are grouped into 8 clusters.  Boosted EMOS-R, EMOS-L and EMOS-C models relying on the extended feature set use training windows of 450, 3283, and 450 calendar days, respectively; for the latter, now 6 clusters are formed. In addition, both SLP and ALP use a further 60-day rolling training period, resulting in a verification period from 2 March to 31 December 2016 for the forecasts investigated in this section.

\begin{table}[t]
  \centering

  {\scriptsize
  \setlength{\tabcolsep}{4pt}

  \begin{tabular}{lccccccc}
    \toprule
    \multicolumn{8}{c}{\textbf{Non-extended feature set}} \\
    \cmidrule(lr){1-8}
    & Raw & EMOS-R & EMOS-C & EMOS-L & DRN & TR & GNN-Geo \\
    \midrule
    Coverage (\%), observed
    & 58.118 & 95.339 & 95.216 & 97.049 & 96.689 & \textbf{95.748} & 96.571 \\
    Width ($^{\circ}$C), observed
    & 3.018 & 7.304 & 7.118 & \textbf{6.820} & 8.201 & 7.666 & 7.615 \\

    \addlinespace[4pt]
    Coverage (\%), unobserved
    & 60.743 & 94.973 & 93.661 & 91.683 & \textbf{96.196} & 95.207 & 95.431 \\
    Width ($^{\circ}$C), unobserved
    & 2.939 & 7.274 & 6.911 & \textbf{6.872} & 8.181 & 7.556 & 7.678 \\
    \bottomrule
  \end{tabular}
  \begin{tabular}{lccccccc}
    \toprule
    \multicolumn{8}{c}{\textbf{Extended feature set}} \\
    \cmidrule(lr){1-8}
    & Raw & EMOS-R & EMOS-C & EMOS-L & DRN & TR & GNN-Geo \\
    \midrule
    Coverage (\%), observed
    & 58.118 & 97.543 & 97.781 & 98.270 & 95.542 & \textbf{95.761} & 94.044 \\
    Width ($^{\circ}$C), observed
    & 3.018 & 7.532 & 7.434 & 7.013 & 5.853 & 5.528 & \textbf{5.280} \\

    \addlinespace[4pt]
    Coverage (\%), unobserved
    & 60.743 & 97.268 & \textbf{96.298} & 92.120 & 92.687 & 92.102 & 88.809 \\
    Width ($^{\circ}$C), unobserved
    & 2.939 & 7.507 & 7.287 & 6.955 & 5.728 & 5.367 & \textbf{5.230} \\
    \bottomrule
  \end{tabular}
  }

  \caption{Coverage and average width of the raw and post-processed 96.08\% prediction intervals for T2M forecasts. Bold values indicate the coverage closest to 96.08\% and the narrowest intervals within each feature set.}
  \label{tab:standalone_coverage_width_t2m}
\end{table}

Table \ref{tab:crpss_table_t2m} presents the CRPSS values of the single (non-pooled) models with respect to the corresponding DRN forecasts for the two feature sets. Note that all post-processed forecasts improve upon the raw ensemble in terms of mean CRPS by a wide margin (see Table \ref{tab:mean_crps_t2m}). For the non-extended feature set, EMOS-L gives the best performance at observed stations, followed by GNN-Geo, EMOS-C, EMOS-R, and the Transformer. The improvements are significant for all these models. For unobserved stations, however, the ranking changes. EMOS-R achieves the highest and significant gain over DRN, and also has the lowest mean CRPS among the competing models (see Table \ref{tab:mean_crps_t2m} in the Appendix). It is followed by the Transformer, which is the only other model with a positive gain over DRN. The remaining models, EMOS-C, GNN-Geo, and EMOS-L, show negative skill in this setting. For the extended feature set, where the standard EMOS framework is replaced by boosted EMOS variants, the machine-learning models become more competitive. At observed stations, the Transformer gives the best CRPSS, followed by GNN-Geo. At unobserved stations, however, only the Transformer remains slightly better than the DRN baseline, and this improvement is not significant. Thus, in terms of mean CRPS, DRN and Transformer can be regarded as the strongest models for spatial interpolation. In contrast, GNN-Geo shows a clear drop from observed to unobserved stations, suggesting that its advantage does not transfer equally well to interpolation sites. One possible explanation is that the geographical graph tends to oversmooth the spatial information. This may be especially relevant for the extended feature set, where the predictors already include information related to spatial variability. In this case, the DRN can use these covariates more directly. This is illustrated by the unobserved Feldberg station in the Black Forest, located at 1,490 m elevation. Since the simple geographical graph does not explicitly account for elevation when selecting neighbouring stations, GNN-Geo performs worse at this location, with a mean CRPS that is substantially higher than that of DRN. Among the EMOS models, when using the extended feature set, EMOS-R performs best at the unobserved stations, while EMOS-L falls clearly behind, which also has the largest generalization gap between observed and unobserved stations among the considered models.

% MAE és RMSE

\begin{figure}
    \centering
    \includegraphics[width=1\linewidth]{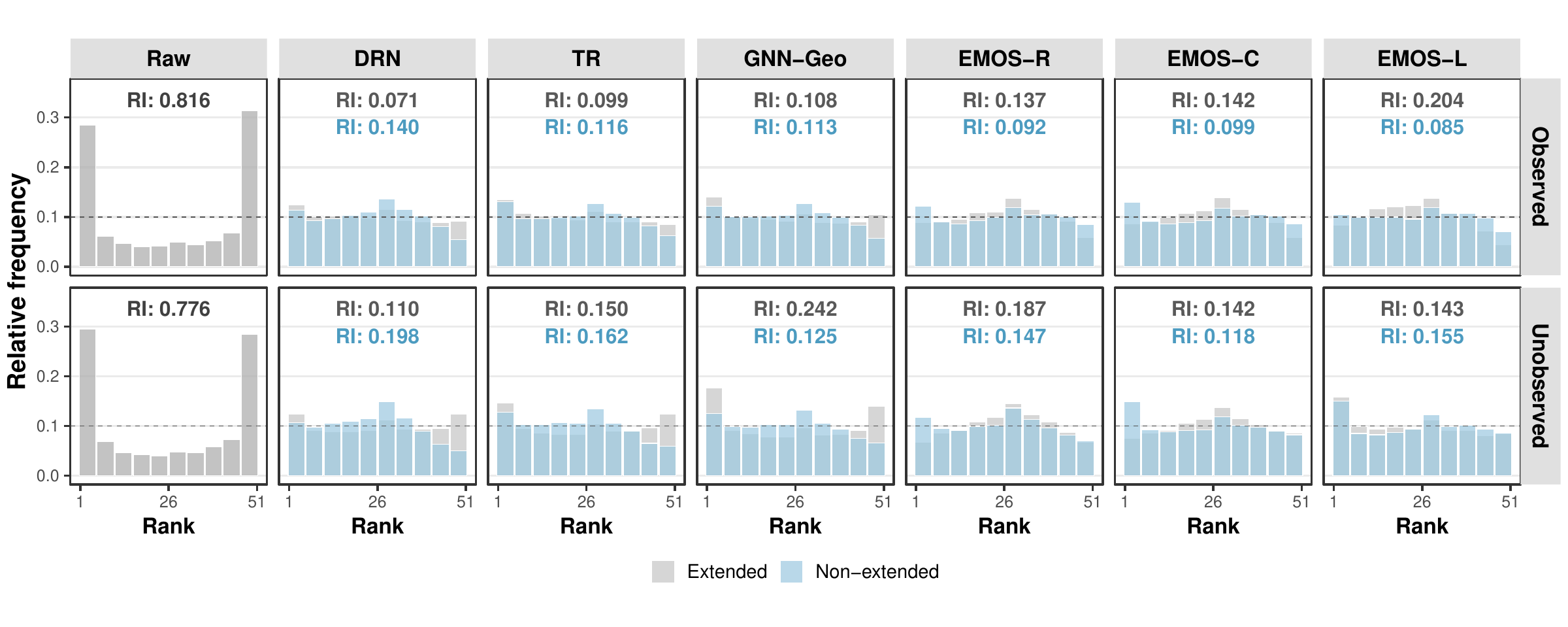}
\caption{Rank histograms and the corresponding RIs of T2M forecasts grouped by feature set and station group.}
    \label{fig:pit_t2m}
\end{figure}

To evaluate the performance of the corresponding point forecasts, Table \ref{tab:mae_rmse_skill_t2m} presents the group-wise mean absolute error skill score (MAES) and root mean squared error skill score (RMSES) for T2M forecasts relative to the corresponding DRN baseline. At observed stations, all post-processed forecasts outperform the raw ensemble, while this is not the case at unobserved sites (see Table \ref{tab:mean_mae_t2m}). For the non-extended feature set, EMOS-L performs best at observed stations in terms of both MAE and RMSE, whereas EMOS-R shows the largest and significant improvement over DRN at unobserved stations. For the extended feature set, the Transformer performs best at observed locations, while GNN-Geo and EMOS-L also show significant improvement. At unobserved locations, however, none of the models significantly outperform DRN. The Transformer has a small positive MAES, but its RMSES is slightly negative. Nevertheless, GNN-Geo performs considerably better at some individual stations. For example, at the unobserved Erfurt-Weimar station, it reduces MAE by 17.45\% and RMSE by 19.10\% relative to DRN (not shown). This station has two observed graph neighbors within 50 km, suggesting that the GNN can benefit from locally available spatial information.

% coverage és avg width

Table \ref{tab:standalone_coverage_width_t2m} presents the coverage and average width of nominal 96.08\% prediction intervals for both feature- and station groups. For the non-extended feature set, among the post-processed forecasts EMOS-L gives the narrowest prediction intervals for both observed and unobserved stations. In terms of coverage, the Transformer is closest to the nominal level at observed stations, while DRN performs best at unobserved stations. For the extended feature set, GNN-Geo produces the narrowest intervals in both station groups, indicating a clear advantage of the AI-based models in terms of interval width. However, the coverage results are more mixed. At observed stations, the Transformer is closest to the nominal level, while at unobserved stations the best coverage is obtained by EMOS-C.

% Rank hisztogram és RI

Figure \ref{fig:pit_t2m} presents the verification rank histograms of the raw and post-processed T2M forecasts, grouped by feature set and station group. At observed stations, all post-processed forecasts show much better calibration than the raw ensemble according to the RI values. For the non-extended feature set, the rank histograms are generally close to uniform, with the lowest RI obtained by EMOS-L, followed by EMOS-R and EMOS-C. For the extended feature set, DRN gives the lowest RI, while the Transformer and GNN-Geo also remain well calibrated. Interestingly, EMOS-L shows the largest increase in RI between observed and unobserved stations for the non-extended feature set. For the extended feature set, the largest increase is observed for GNN-Geo. At unobserved stations, DRN has the lowest RI for the extended feature set, whereas EMOS-C performs best for the non-extended feature set.

% LP

\begin{table}[!t]
  \centering

  {\scriptsize
  \setlength{\tabcolsep}{4pt}
  \begin{tabular}{lccc@{\hspace{12pt}}lccc}
    \toprule
    \multicolumn{4}{c}{\textbf{Observed stations}
      \quad (EMOS-L: 0.7823)}
    &
    \multicolumn{4}{c}{\textbf{Unobserved stations}
      \quad (EMOS-L: 1.0656)} \\
    \cmidrule(lr){1-4}
    \cmidrule(lr){5-8}

    ML model & Standalone & SLP & ALP
    & ML model & Standalone & SLP & ALP \\
    \midrule

    DRN
      & 0.7898 & 0.7554 & \textbf{0.7552}
    & DRN
      & \textbf{0.8743} & 0.9051 & 0.8812$^{*}$ \\

    TR
      & 0.7595 & 0.7341 & \textbf{0.7338}
    & TR
      & 0.8689 & 0.8708 & \textbf{0.8613$^{*}$} \\

    GNN-Geo
      & 0.7688 & 0.7400 & \textbf{0.7394}
    & GNN-Geo
      & 0.9450 & 0.9222 & \textbf{0.9092$^{*}$} \\

    \bottomrule
  \end{tabular}
  }

  \caption{Mean CRPS of EMOS-L, the selected machine-learning forecasts in the case of the exntended feature set
  and their two-component linear pools. Each pool combines EMOS-L with the
  model shown in the corresponding ML model column. Lower values indicate
  better predictive performance, while bold values mark the best result
  within each row and station group. Asterisks indicate that ALP
  significantly outperforms the corresponding SLP according to one-sided
  Diebold--Mariano tests at the 5\% level.}
  \label{tab:linear_pool_crps_components_t2m}
\end{table}

As mentioned in Section \ref{subs3.3}, for the T2M forecasts we also consider linear pools (LP) of competing forecasts. Here, we focus on the combination of the best statistical competitor and the ML-based models using the extended feature set. Thus, we investigate three combinations: EMOS-L + DRN, EMOS-L + TR, and EMOS-L + GNN-Geo. As mentioned, as a baseline, we use the SLP of \citet{baran2018combining}, where the component weights are estimated in a rolling training window by minimizing the two-component mixture CRPS on historical data. In contrast to \citet{baran2018combining}, the weights in our setting are estimated using only the observed stations, and the same weights are then applied to the unobserved locations. Moreover, we also consider the ALP, also discussed in detail in Section \ref{subs3.3}. 

Table \ref{tab:linear_pool_crps_components_t2m} presents the mean CRPS values for the individual standalone forecasts, and for the SLP and ALP predictions. For observed stations, both linear pools improve upon the individual components, but there is no relevant difference between the two pooling methods. This suggests that, at observed stations, the global rolling pool is already sufficient to improve the forecasts in terms of mean CRPS. For unobserved stations, however, the picture is different. The SLP is not always better than the best individual component, indicating that the combination can sometimes weaken the useful information of the individual forecasts. In contrast, the ALP is significantly better than the global rolling pool in all three cases, achieves the lowest mean CRPS, and also outperforms the individual components.

\begin{table}[t]
  \centering

  {\scriptsize
  \setlength{\tabcolsep}{5pt}

  \begin{tabular}{lccccc}
    \toprule
    \multicolumn{6}{c}{\textbf{Non-extended feature set}} \\
    \cmidrule(lr){1-6}
    & EMOS-R & EMOS-C & EMOS-L & TR & GNN-Geo \\
    \midrule
    Observed
    & -1.225 & 3.052$^{*}$ & \textbf{26.762$^{*}$}
    & -1.413 & 5.337$^{*}$ \\
    Unobserved
    & 0.012 & \textbf{0.679$^{*}$} & -27.113
    & -1.816 & -0.206 \\
    \bottomrule
  \end{tabular}
  \begin{tabular}{lccccc}
    \toprule
    \multicolumn{6}{c}{\textbf{Extended feature set}} \\
    \cmidrule(lr){1-6}
    & EMOS-R & EMOS-C & EMOS-L & TR & GNN-Geo \\
    \midrule
    Observed
    & -33.739 & -27.262 & 1.553$^{*}$
    & \textbf{4.892$^{*}$} & -11.263 \\
    Unobserved
    & 6.899$^{*}$ & \textbf{7.070$^{*}$} & -45.230
    & -1.699 & 4.854$^{*}$ \\
    \bottomrule
  \end{tabular}
  }

  \caption{Group-wise CRPS skill scores of WS forecasts relative to the
  corresponding DRN baseline.
  Positive values indicate improvement, while bold values mark the best
  model within each row and feature set. Asterisks indicate significant
  improvement according to one-sided Diebold--Mariano tests at the 5\% level.}
  \label{tab:crpss_table_ws}
\end{table}

\subsection{Wind-speed}
\label{sec4.2}

In this subsection, we present the results for WS forecasts using 2016 as the verification period. The rolling training period is 3283 days for all machine-learning models, except for the GNN-Geo model, where we used 361 days. For the classical EMOS-R, EMOS-L, and EMOS-C models using the non-extended feature set, the training periods are 35, 60, and 90 days, respectively. For EMOS-C, the observed stations are divided into 10 clusters. The boosted EMOS-R, EMOS-L, and EMOS-C models using the extended feature set have training periods of 70, 3283, and 450 days, respectively. For the boosted EMOS-C model, the observed stations are divided into 4 clusters.

\begin{table}[t]
  \centering
  {\scriptsize
  \setlength{\tabcolsep}{5pt}
  \begin{tabular}{lccccc}
    \toprule
    \multicolumn{6}{c}{\textbf{Non-extended feature set}} \\
    \cmidrule(lr){1-6}
    & EMOS-R & EMOS-C & EMOS-L & TR & GNN-Geo \\
    \midrule
    MAES, observed & -0.641 & -0.319 & \textbf{25.938$^{*}$} & -1.874 & 4.782$^{*}$ \\
    MAES, unobserved & \textbf{0.373} & -0.821 & -23.073 & -2.051 & 0.251 \\
    \addlinespace[3pt]
    RMSES, observed & -1.320 & -0.189 & \textbf{31.070$^{*}$} & -0.430 & 5.159$^{*}$ \\
    RMSES, unobserved & -0.103 & -0.815 & -32.219 & -1.253 & -0.105 \\
    \bottomrule
  \end{tabular}
  \vspace{0.7em}
  \begin{tabular}{lccccc}
    \toprule
    \multicolumn{6}{c}{\textbf{Extended feature set}} \\
    \cmidrule(lr){1-6}
    & EMOS-R & EMOS-C & EMOS-L & TR & GNN-Geo \\
    \midrule
    MAES, observed & -31.603 & -25.037 & 2.015$^{*}$ & \textbf{4.533$^{*}$} & -11.250 \\
    MAES, unobserved & 4.868$^{*}$ & \textbf{5.829$^{*}$} & -42.502 & -0.192 & 3.184$^{*}$ \\
    \addlinespace[3pt]
    RMSES, observed & -37.674 & -25.172 & 1.771$^{*}$ & \textbf{3.578$^{*}$} & -11.537 \\
    RMSES, unobserved & 4.918$^{*}$ & \textbf{6.428$^{*}$} & -59.880 & 0.475 & 5.442$^{*}$ \\
    \bottomrule
  \end{tabular}
  }
  \caption{Group-wise mean absolute error skill scores (MAES) and root mean squared error skill scores (RMSES) of WS forecasts relative to the corresponding DRN baseline. Positive values indicate improvement, while bold values mark the best model within each row and feature set. Asterisks indicate significant improvement according to one-sided Diebold--Mariano tests at the 5\% level.}
  \label{tab:mae_rmse_skill_ws}
\end{table}

Table \ref{tab:crpss_table_ws} presents the CRPSS values of WS forecasts relative to the corresponding DRN baseline. Most post-processed forecasts outperform the raw ensemble (see Table \ref{tab:mean_crps_ws_appendix}), although EMOS-L performs worse at unobserved stations. For the non-extended feature set, EMOS-L provides the largest and significant improvement over DRN at observed stations, while EMOS-C shows a small but significant improvement at the unobserved ones. Similar to the T2M results, EMOS-L has a substantial generalization gap. With the extended feature set, the Transformer provides the largest improvement over DRN at observed stations. At unobserved stations, EMOS-C performs best, followed closely by EMOS-R, while GNN-Geo also significantly outperforms DRN. An exploratory analysis shows that the relative performance of EMOS-C and DRN depends on the forecast situation. With the non-extended feature set, EMOS-C performs better on 59.2\% of the evaluated days. This proportion increases to 86.0\% with the extended feature set. DRN performs better for medium wind speeds and when the raw ensemble error is small. In contrast, EMOS-C performs particularly well in more difficult situations with larger raw ensemble errors. This suggests that boosting is especially useful in these cases because it can select additional predictors that describe more complex forecast situations.

Table \ref{tab:mae_rmse_skill_ws} reports the MAES and RMSES values for the competing WS forecasts, using the DRNs as baseline models. In the case of the non-extended feature set, EMOS-L clearly performs best at the observed stations in terms of both MAES and RMSES. However, this advantage of the local model disappears at the unobserved stations, where EMOS-L performs worse, which indicates that EMOS-L can not successfully transfer to unobserved locations. For the observed stations, GNN-Geo performs closest to EMOS-L, with a significantly positive skill and a substantially smaller generalization gap. Based on the MAES values of unobserved stations, EMOS-R has a moderate but significant skill over the DRN, closely followed by the GNN-Geo, while none of the models perform better than the DRN based on RMSES. In the case of the extended feature set, however, the Transformer performs best at the observed stations in terms of both MAES and RMSES, while at the unobserved stations, EMOS-C shows the highest gain over the DRN, closely followed by EMOS-R and GNN-Geo. Note that, based on the MAE and RMSE values for forecasts at unobserved locations, not all models perform better than the raw ensemble (see Tables \ref{tab:mean_mae_ws} and \ref{tab:mean_rmse_ws}).

Table \ref{tab:standalone_coverage_width_ws} presents the coverage and average width of 96.08\% prediction intervals for WS forecasts. In the case of the non-extended feature set, GNN-Geo has the coverage closest to the nominal level at the observed stations, while DRN performs best at the unobserved stations. However, among the post-processed predictions, EMOS-L provides the narrowest prediction intervals for both observed and unobserved stations. While the Transformer provides the narrowest prediction intervals with the extended feature set, the results are more mixed in terms of coverage. DRN has the coverage closest to the nominal level at the observed stations, while EMOS-C performs best at the unobserved stations, closely followed by EMOS-R.

\begin{figure}
    \centering
    \includegraphics[width=1\linewidth]{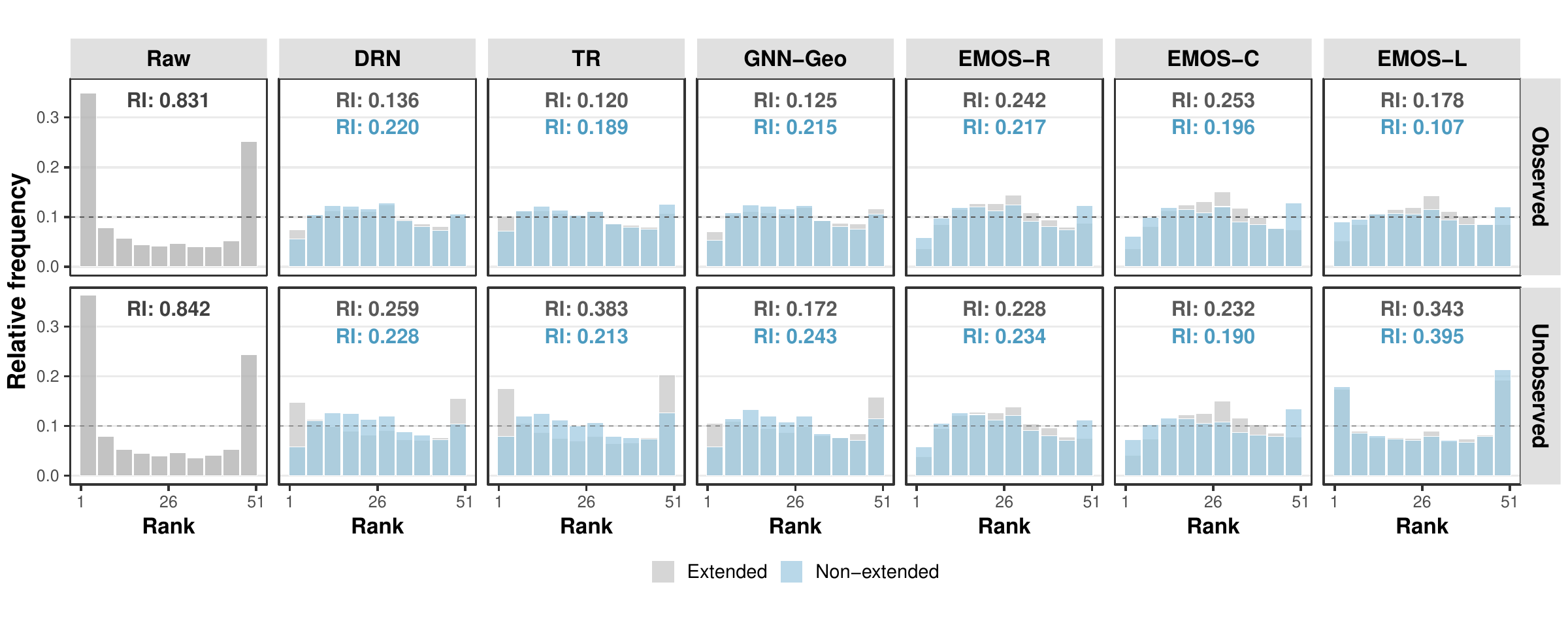}
\caption{Rank histograms and the corresponding RIs of WS forecasts grouped by feature set and station group.}
    \label{fig:pit_ws}
\end{figure}

Finally, Figure \ref{fig:pit_ws} presents the rank histograms and the corresponding RIs of WS forecasts grouped by feature set and station group. Similar to the T2M predictions, all post-processed predictions outperform the raw ensemble forecasts. In the case of the observed stations, EMOS-L has the best calibration based on the RI, followed by the Transformer and closely by EMOS-C when using the non-extended feature set. However, when using more predictors, the ranking changes. The Transformer has the lowest RI, closely followed by the other two machine learning models. In the case of the unobserved stations, EMOS-C has the lowest RI when using the limited feature set, followed by the Transformer and DRN. However, when using more predictors, GNN-Geo obtains the lowest RI, followed by EMOS-R and EMOS-C.

%%%%
For wind speed, the CRPS of a linear pool of truncated normal distributions is not available in closed form. Although it can be calculated numerically, this requires additional approximations and computational effort. Therefore, we do not consider linear pools for wind speed.

\begin{table}[t]
  \centering

  {\scriptsize
  \setlength{\tabcolsep}{4pt}

  \begin{tabular}{lccccccc}
    \toprule
    \multicolumn{8}{c}{\textbf{Non-extended feature set}} \\
    \cmidrule(lr){1-8}
    & Raw & EMOS-R & EMOS-C & EMOS-L & DRN & TR & GNN-Geo \\
    \midrule
    Coverage (\%), observed
    & 59.311 & 93.666 & 93.567 & 94.291 & 95.429 & 94.380 & \textbf{95.785} \\
    Width (m/s), observed
    & 2.433 & 5.616 & 5.698 & \textbf{4.485} & 6.038 & 5.650 & 5.800 \\

    \addlinespace[4pt]
    Coverage (\%), unobserved
    & 58.981 & 94.484 & 92.952 & 81.712 & \textbf{95.401} & 94.002 & 94.627 \\
    Width (m/s), unobserved
    & 2.343 & 5.542 & 5.255 & \textbf{4.364} & 5.752 & 5.304 & 5.530 \\
    \bottomrule
  \end{tabular}
\begin{tabular}{lccccccc}
    \toprule
    \multicolumn{8}{c}{\textbf{Extended feature set}} \\
    \cmidrule(lr){1-8}
    & Raw & EMOS-R & EMOS-C & EMOS-L & DRN & TR & GNN-Geo \\
    \midrule
    Coverage (\%), observed
    & 59.311 & 97.139 & 97.945 & 97.592 & \textbf{96.296} & 95.370 & 95.840 \\
    Width (m/s), observed
    & 2.433 & 6.577 & 6.894 & 4.982 & 4.560 & \textbf{4.090} & 4.873 \\

    \addlinespace[4pt]
    Coverage (\%), unobserved
    & 58.981 & 97.968 & \textbf{97.907} & 84.350 & 90.103 & 85.108 & 92.566 \\
    Width (m/s), unobserved
    & 2.343 & 6.306 & 6.479 & 4.956 & 4.332 & \textbf{3.857} & 4.634 \\
    \bottomrule
  \end{tabular}
  }

  \caption{Coverage and average width of the raw and post-processed 96.08\% prediction intervals for WS forecasts. Bold values indicate the coverage closest to 96.08\% and the narrowest intervals within each feature set.}
  \label{tab:standalone_coverage_width_ws}
\end{table}

\section{Discussion}

This study provides a comprehensive comparison of different statistical and machine-learning-based post-processing techniques for obtaining calibrated predictions at locations where the observations, which are crucial training targets, are not available. For all models, we investigated two different setups. In the first setup, we only used predictors that are typically included in the standard EMOS framework, namely the mean and standard deviation of the raw ensemble forecasts. In the second setup, we used all predictors listed in Table \ref{tab:predictors} to assess how the amount of input information affects the performance of the post-processing models. Hence, for the post-processing of T2M and WS forecasts, we applied the standard EMOS models when using the non-extended feature set, while their boosted variants replaced them when using the extended feature set. As machine learning models, we use DRNs, Transformers, and GNNs, as they represent different approaches to learning the relationship between the input features and the target variable. Moreover, in the case of T2M predictions, we also investigated the linear combination of the best-performing EMOS model with the individual machine learning models and proposed a simple extension of the SLP for unobserved locations. Furthermore, we proposed an altitude-aware variant of the SLP to include additional location-specific information in the combination.

In general, we found that all post-processed forecasts outperform the raw ensemble forecasts in most aspects, but there is no universal approach that performs best in all settings and according to considered metrics. For T2M, the regional models generally performed well, while more sophisticated approaches, such as the method proposed by \cite{baran2024clustering}, did not show a clear advantage. However, for WS predictions, the semi-local EMOS clearly emerged as one of the best-performing models. This suggests that in more difficult and heterogeneous settings, more sophisticated approaches can be advantageous for spatial interpolation. Based on the results, machine learning models could be a promising alternative for providing forecasts at unobserved locations, even when using the limited feature set. This is especially true for T2M, where the Transformer showed a good compromise across the considered evaluation metrics. An important aspect of the results is the difference in performance between observed and unobserved stations. In general, good performance at observed stations does not necessarily guarantee good performance at locations without observations. This is especially visible for local models, which can strongly benefit from station-specific information but may have difficulties when transferred to new locations. For example, EMOS-L performs very well at observed stations, while its performance decreases considerably at unobserved stations. In contrast, regional, semi-local, and machine learning models generally show a better ability to transfer information to new locations.

These differences between the models also motivate the use of forecast combinations, as different approaches may capture different aspects of the relationship between the forecasts and observations. For T2M, the proposed ALP achieves the best performance for all three model combinations at observed stations. At unobserved stations, the combination faces a more difficult task due to the large generalization gap of EMOS-L. Nevertheless, the ALP achieves the lowest CRPS in two out of three cases and significantly improves over the corresponding SLP in all three cases. As temperature is strongly related to altitude, these results suggest that including altitude information can help improve the transfer of combined forecasts to unobserved locations.

Future work could focus on further improving the spatial modeling of the proposed approaches. For GNN-Geo, more sophisticated graph structures could be investigated to better represent the relationships between stations, for example by including additional geographical or meteorological information when defining the connections. Furthermore, the proposed ALP could be extended by incorporating additional location-specific information beyond altitude, which may further improve its performance at unobserved locations.

\section*{Acknowledgments}
\label{ack}
The research was supported by the Hungarian National Research, Development and Innovation Office under Grant K142849 and by the EKÖP-26-4 University Research Scholarship Programme of the Ministry for Science and Technology, funded by the National Research, Development and Innovation Fund.

The author would like to thank Sándor Baran for performing the boosted EMOS analysis and for providing comments on an earlier version of the manuscript.
\bibliographystyle{rss}

\bibliography{SpatialInt_paper}

\appendix
\renewcommand{\thetable}{A\arabic{table}}
\renewcommand{\thefigure}{A\arabic{figure}}
\setcounter{table}{0}
\setcounter{figure}{0}
\section{Additional tables}

This section provides additional information about the predictors and hyperparameter settings used in the study. It also reports the mean CRPS, MAE, and RMSE of the raw and post-processed temperature and wind speed forecasts, separately for observed and unobserved stations and for the two feature sets. These tables complement the skill scores and comparisons presented in Section \ref{sec4}.

\begin{table}[ht]
\centering
\small
\begin{tabular}{p{0.22\linewidth} p{0.58\linewidth} cc}
\toprule
\textbf{Variable} & \textbf{Description} & \textbf{T2M} & \textbf{WS} \\
\midrule

\multicolumn{4}{l}{\textit{Meteorological variables}} \\
\texttt{t2m}        & 2-m temperature  & \cmark & \cmark \\
\texttt{ws}         & 10-m wind speed      & \xmark & \cmark \\
\texttt{d2m}        & 2-m dewpoint temperature                   & \cmark & \cmark \\
\texttt{cape}       & Convective available potential energy      & \cmark & \cmark \\
\texttt{sp}         & Surface pressure                           & \cmark & \cmark \\
\texttt{tcc}        & Total cloud cover                          & \cmark & \cmark \\
\texttt{u10}        & 10-m zonal wind component                  & \cmark & \cmark \\
\texttt{v10}        & 10-m meridional wind component             & \cmark & \cmark \\
\texttt{u\_pl850}   & Zonal wind component at 850 hPa            & \cmark & \cmark \\
\texttt{v\_pl850}   & Meridional wind component at 850 hPa       & \cmark & \cmark \\
\texttt{ws\_pl850}  & Wind speed at 850 hPa                      & \xmark & \cmark \\
\texttt{q\_pl850}   & Specific humidity at 850 hPa               & \cmark & \cmark \\
\texttt{u\_pl500}   & Zonal wind component at 500 hPa            & \cmark & \cmark \\
\texttt{v\_pl500}   & Meridional wind component at 500 hPa       & \cmark & \cmark \\
\texttt{ws\_pl500}  & Wind speed at 500 hPa                      & \xmark & \cmark \\
\texttt{gh\_pl500}  & Geopotential height at 500 hPa             & \cmark & \cmark \\
\texttt{sshf}       & Sensible heat flux                         & \cmark & \cmark \\
\texttt{slhf}       & Latent heat flux                           & \cmark & \cmark \\
\texttt{ssr}        & Surface shortwave radiation flux           & \cmark & \cmark \\
\texttt{str}        & Surface longwave radiation flux            & \cmark & \cmark \\
\midrule

\multicolumn{4}{l}{\textit{Station metadata}} \\
\texttt{lat}        & Station latitude                           & \cmark & \cmark \\
\texttt{lon}        & Station longitude                          & \cmark & \cmark \\
\texttt{alt}        & Station altitude                           & \cmark & \cmark \\
\texttt{orog}       & Model grid-point altitude (orography)      & \cmark & \xmark \\
\midrule

\multicolumn{4}{l}{\textit{Temporal variables}} \\
\texttt{sin\_yday}  & Sine-transformed day of year               & \cmark & \cmark \\
\texttt{yday}       & Day of year                                & \cmark & \cmark \\
\texttt{month}      & Month of year                              & \cmark & \cmark \\
\bottomrule
\end{tabular}
\caption{Variables available in the dataset. For meteorological variables, ensemble information is summarized by the ensemble mean and standard deviation, both of which are included as predictors whenever the corresponding variable is marked. Columns \textbf{T2M} and \textbf{WS} indicate whether a variable is used in the respective post-processing model.}
\label{tab:predictors}
\end{table}

\begin{table}[t]
\scriptsize
\setlength{\tabcolsep}{5pt}
\centering
\begin{tabular}{llrrrrrl}
\toprule
Model & Feature set & Batch size & Learning rate & Hidden dim. & Layers & Dropout & Heads \\
\midrule
\multicolumn{8}{c}{\textbf{T2M}} \\
\midrule
GNN & Both & 256 & 0.02 & 256 & 1 & 0.2 & -- \\
MLP & Non-extended & 1200 & 0.01 & 64 & 1 & 0.2 & -- \\
MLP & Extended & 4800 & 0.01 & 256 & 1 & 0.3 & -- \\
Transformer & Both & 64 & 0.01 & 64 & 1 & 0.1 & 1 head \\
\midrule
\multicolumn{8}{c}{\textbf{WS}} \\
\midrule
GNN & Both & 64 & 0.01 & 32 & 1 & 0.1 & -- \\
MLP & Both & 9600 & 0.02 & 256 & 1 & 0.2 & -- \\
Transformer & Both & 64 & 0.003 & 256 & 1 & 0.1 & 4 heads \\
\bottomrule
\end{tabular}
\caption{Selected hyperparameter settings for the machine-learning models.}
\label{tab:ml_hyperparameters}
\end{table}

\begin{table}[t]
  \centering
  \small
  \setlength{\tabcolsep}{3pt}

  \begin{adjustbox}{max width=\linewidth}
    \begin{tabular}{lccccccccccccc}
      \toprule
      & Raw
      & \multicolumn{6}{c}{\textbf{Non-extended feature set}}
      & \multicolumn{6}{c}{\textbf{Extended feature set}} \\
      \cmidrule(lr){3-8}
      \cmidrule(lr){9-14}
      & -- & EMOS-R & EMOS-C & EMOS-L & DRN & TR & GNN-Geo
      & EMOS-R & EMOS-C & EMOS-L & DRN & TR & GNN-Geo \\
      \midrule
      Observed
      & 1.141 & 0.992 & 0.973 & \textbf{0.858} & 1.006 & 1.002 & 0.962
      & 0.894 & 0.866 & 0.782 & 0.790 & \textbf{0.760} & 0.769 \\
      Unobserved
      & 1.102 & \textbf{0.971} & 0.998 & 1.040 & 0.996 & 0.989 & 1.015
      & 0.889 & 0.919 & 1.066 & 0.874 & \textbf{0.869} & 0.945 \\
      \bottomrule
    \end{tabular}
  \end{adjustbox}

  \caption{Mean CRPS of the raw- and post-processed T2M forecasts.
  Lower values indicate better predictive performance, while bold values
  indicate the best performance within each feature set.}
  \label{tab:mean_crps_t2m}
\end{table}

\begin{table}[t]
  \centering
  \small
  \setlength{\tabcolsep}{3pt}

  \begin{adjustbox}{max width=\linewidth}
    \begin{tabular}{lccccccccccccc}
      \toprule
      & Raw
      & \multicolumn{6}{c}{\textbf{Non-extended feature set}}
      & \multicolumn{6}{c}{\textbf{Extended feature set}} \\
      \cmidrule(lr){3-8}
      \cmidrule(lr){9-14}
      & -- & EMOS-R & EMOS-C & EMOS-L & DRN & TR & GNN-Geo & EMOS-R & EMOS-C & EMOS-L & DRN & TR & GNN-Geo \\
      \midrule
      Observed & 1.416 & 1.385 & 1.364 & \textbf{1.209} & 1.396 & 1.394 & 1.344 & 1.247 & 1.208 & 1.088 & 1.106 & \textbf{1.071} & 1.075 \\
      Unobserved & 1.371 & \textbf{1.333} & 1.379 & 1.412 & 1.362 & 1.362 & 1.404 & 1.221 & 1.261 & 1.463 & 1.203 & \textbf{1.202} & 1.288 \\
      \bottomrule
    \end{tabular}
  \end{adjustbox}

  \caption{Mean absolute error (MAE) of the raw- and post-processed T2M forecasts. Lower values indicate better predictive performance, while bold values indicate the best performance within each feature set.}
  \label{tab:mean_mae_t2m}
\end{table}

\begin{table}[t]
  \centering
  \small
  \setlength{\tabcolsep}{3pt}

  \begin{adjustbox}{max width=\linewidth}
    \begin{tabular}{lccccccccccccc}
      \toprule
      & Raw
      & \multicolumn{6}{c}{\textbf{Non-extended feature set}}
      & \multicolumn{6}{c}{\textbf{Extended feature set}} \\
      \cmidrule(lr){3-8}
      \cmidrule(lr){9-14}
      & -- & EMOS-R & EMOS-C & EMOS-L & DRN & TR & GNN-Geo & EMOS-R & EMOS-C & EMOS-L & DRN & TR & GNN-Geo \\
      \midrule
      Observed & 1.817 & 1.782 & 1.758 & \textbf{1.560} & 1.811 & 1.810 & 1.739 & 1.620 & 1.564 & 1.404 & 1.428 & \textbf{1.386} & 1.389 \\
      Unobserved & 1.815 & \textbf{1.782} & 1.831 & 1.909 & 1.816 & 1.815 & 1.873 & 1.666 & 1.731 & 1.951 & \textbf{1.624} & 1.626 & 1.736 \\
      \bottomrule
    \end{tabular}
  \end{adjustbox}

  \caption{Root mean squared error (RMSE) of the raw- and post-processed T2M forecasts. Lower values indicate better predictive performance, while bold values indicate the best performance within each feature set.}
  \label{tab:rmse_t2m}
\end{table}

\begin{table}[t]
  \centering
  \small
  \setlength{\tabcolsep}{3pt}

  \begin{adjustbox}{max width=\linewidth}
    \begin{tabular}{lccccccccccccc}
      \toprule
      & Raw
      & \multicolumn{6}{c}{\textbf{Non-extended feature set}}
      & \multicolumn{6}{c}{\textbf{Extended feature set}} \\
      \cmidrule(lr){3-8}
      \cmidrule(lr){9-14}
      & -- & EMOS-R & EMOS-C & EMOS-L & DRN & TR & GNN-Geo
      & EMOS-R & EMOS-C & EMOS-L & DRN & TR & GNN-Geo \\
      \midrule
    Observed & 0.986 & 0.865 & 0.828 & \textbf{0.625} & 0.855 & 0.867 & 0.810 & 0.806 & 0.768 & 0.595 & 0.604 & \textbf{0.574} & 0.672 \\
    Unobserved & 0.966 & 0.841 & \textbf{0.835} & 1.074 & 0.841 & 0.856 & 0.843 & 0.769 & \textbf{0.768} & 1.204 & 0.827 & 0.840 & 0.786 \\
      \bottomrule
    \end{tabular}
  \end{adjustbox}

  \caption{Mean CRPS of the raw- and post-processed WS forecasts.
  Lower values indicate better predictive performance, while bold values
  indicate the best performance within each feature set.}
  \label{tab:mean_crps_ws_appendix}
\end{table}

\begin{table}[t]
  \centering
  \small
  \setlength{\tabcolsep}{3pt}

  \begin{adjustbox}{max width=\linewidth}
    \begin{tabular}{lccccccccccccc}
      \toprule
      & Raw
      & \multicolumn{6}{c}{\textbf{Non-extended feature set}}
      & \multicolumn{6}{c}{\textbf{Extended feature set}} \\
      \cmidrule(lr){3-8}
      \cmidrule(lr){9-14}
      & -- & EMOS-R & EMOS-C & EMOS-L & DRN & TR & GNN-Geo & EMOS-R & EMOS-C & EMOS-L & DRN & TR & GNN-Geo \\
      \midrule
      Observed & 1.230 & 1.198 & 1.194 & \textbf{0.882} & 1.190 & 1.213 & 1.134 & 1.117 & 1.061 & 0.831 & 0.848 & \textbf{0.810} & 0.944 \\
      Unobserved & 1.183 & \textbf{1.153} & 1.167 & 1.426 & 1.158 & 1.181 & 1.155 & 1.075 & \textbf{1.064} & 1.610 & 1.130 & 1.132 & 1.094 \\
      \bottomrule
    \end{tabular}
  \end{adjustbox}

  \caption{Mean absolute error (MAE) of the raw- and post-processed WS forecasts. Lower
values indicate better predictive performance, while bold values indicate the best perfor-
mance within each feature set.}
  \label{tab:mean_mae_ws}
\end{table}

\begin{table}[t]
  \centering
  \small
  \setlength{\tabcolsep}{3pt}

  \begin{adjustbox}{max width=\linewidth}
    \begin{tabular}{lccccccccccccc}
      \toprule
      & Raw
      & \multicolumn{6}{c}{\textbf{Non-extended feature set}}
      & \multicolumn{6}{c}{\textbf{Extended feature set}} \\
      \cmidrule(lr){3-8}
      \cmidrule(lr){9-14}
      & -- & EMOS-R & EMOS-C & EMOS-L & DRN & TR & GNN-Geo & EMOS-R & EMOS-C & EMOS-L & DRN & TR & GNN-Geo \\
      \midrule
      Observed & 1.836 & 1.822 & 1.801 & \textbf{1.239} & 1.798 & 1.806 & 1.705 & 1.620 & 1.473 & 1.156 & 1.176 & \textbf{1.134} & 1.312 \\
      Unobserved & 1.747 & 1.742 & 1.754 & 2.301 & \textbf{1.740} & 1.762 & 1.742 & 1.533 & \textbf{1.509} & 2.578 & 1.612 & 1.605 & 1.525 \\
      \bottomrule
    \end{tabular}
  \end{adjustbox}

  \caption{Root mean squared error (RMSE) of the raw- and post-processed WS forecasts.
Lower values indicate better predictive performance, while bold values indicate the best
performance within each feature set}
  \label{tab:mean_rmse_ws}
\end{table}

\end{document}